\documentclass[letterpaper]{article} 
\usepackage[preprint]{aaai2027}  
\usepackage[hyphens]{url}  
\usepackage{graphicx} 
\usepackage{natbib}  
\usepackage{caption} 
\usepackage{booktabs}
\usepackage{amsmath}
\usepackage{amssymb}

\title{EcoAgent-Bench: Evaluating Economic Decision-Making in Budget-Constrained LLM Agents}
\author{
    Jie Wu, Ming Gong, Feixiang Cheng, Qinqin Zhao
}
\affiliations{
    Atlassian
}

\begin{document}
\maketitle

\begin{abstract}
Agent benchmarks usually measure task completion and treat resource use as an auxiliary statistic. In deployment, however, the choice among a local lookup, broad search, composite research tool, stronger model, or human escalation is part of the task itself. We introduce \textbf{EcoAgent-Bench}, in which every task specifies priced actions and an explicit budget. Its 304 real-derived tasks span five families adapted from GAIA, HotpotQA, and MuSiQue, and test four decisions: avoiding unnecessary escalation, escalating when local evidence is insufficient, selecting a model tier, and stopping on unsupported premises. We evaluate seven LLM agents in tool-API and workspace-CLI settings, together with four oracle scripted controls. Micro-averaged accuracy rewards one-sided policies: always-escalate controls achieve high micro success while failing save-oriented tasks. We therefore also report an \emph{economic-consistency} score---the worse of accuracy on upgrade-oriented and save-oriented family groups---which exposes this failure. Tool-API agents attain only 3.9--24.0\% micro strict success (at most 7.3\% economic consistency), often either stopping before warranted escalation or overspending on cheap tasks. A threshold-crossing budget sweep changes GPT-5.4's escalation rate from 0\% to only 3\%. These results show that completion under a budget and economical action selection are distinct properties. We release the task bundle, transformation pipeline, frozen evaluation environments, and integrity-bound result artifacts needed to study both.
\end{abstract}

\section{Introduction}

LLM-based agents browse the web, write code, query databases, and orchestrate multi-step workflows.
Every action, however, has a cost: API calls consume tokens, composite tools charge premiums, and human escalation is expensive.
Existing agent benchmarks---GAIA~\cite{mialon2023gaia}, SWE-bench~\cite{jimenez2023swebench}, WebArena~\cite{zhou2023webarena}, AgentBench~\cite{liu2023agentbench}, and $\tau$-bench~\cite{yao2024taubench}---primarily measure task completion; resource use is usually reported after the episode.
Few benchmarks make a stated budget and priced actions part of the task instance, such that the preferred action changes with the available evidence.

We argue this is a critical gap.
An agent that always calls the most expensive tool may be accurate but wasteful; an agent that always uses the cheapest tool may be frugal but wrong when the cheap evidence is insufficient.
The economically rational agent makes \emph{conditional} decisions: use cheap evidence when it suffices, escalate only when the evidence warrants it, route to a stronger (pricier) model only when the reasoning demands it, and stop---rather than keep spending---when no answer can be found.

\begin{figure*}[t]
\centering
\includegraphics[width=\textwidth]{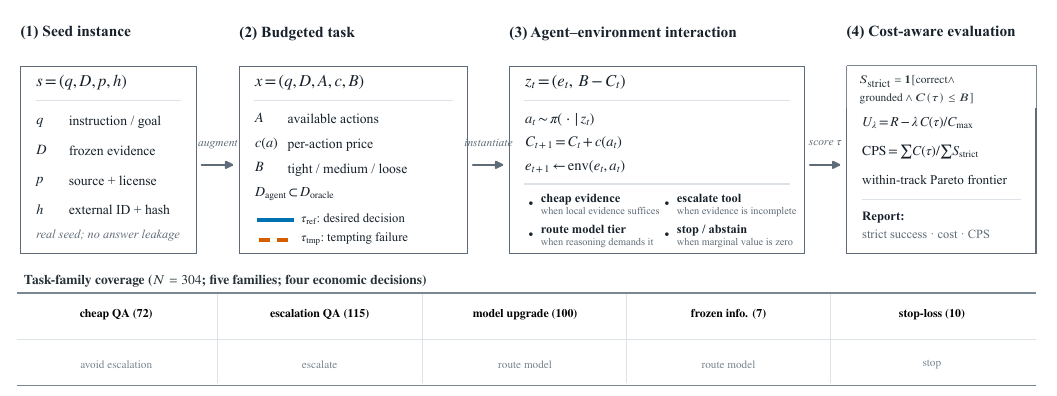}
\caption{EcoAgent-Bench pipeline. Dataset seeds are transformed into frozen tasks with priced actions, separate agent/evaluator views, and canonical and high-regret traces. Evaluation records correctness, evidence access, budget feasibility, and trajectory cost across five task families.}
\label{fig:overview}
\end{figure*}

\textbf{EcoAgent-Bench} operationalizes this setting (Figure~\ref{fig:overview}).
Tasks are derived from established datasets rather than written solely to illustrate a cost trade-off.
The transformation retains the source question or issue while adding a budgeted tool environment, a canonical policy trace, a contrasting high-regret trace, and separate agent and evaluator views.

\textbf{Contributions.}
(1)~A budget-conditioned task format in which cost is part of the task, not a logging afterthought.
(2)~A 304-task, five-family real-derived bundle covering four economic decisions.
(3)~A reproducible transformation protocol with deterministic selection, leakage and provenance checks, offline model-tier labels, verified false-premise stop tasks, and integrity-bound result artifacts.
(4)~An evaluation of seven LLM agents plus four oracle controls across three tracks, including a threshold-crossing budget sweep that tests whether tool selection responds to the available budget.
(5)~An \emph{economic-consistency} diagnostic---the minimum of accuracy on upgrade-oriented and save-oriented family groups---that exposes one-sided policies obscured by plain micro-averaged accuracy.

\section{Related Work}

\textbf{Agent benchmarks.}
GAIA~\cite{mialon2023gaia}, SWE-bench~\cite{jimenez2023swebench}, WebArena~\cite{zhou2023webarena}, and AgentBench~\cite{liu2023agentbench} evaluate whether agents can \emph{complete} tasks; $\tau$-bench~\cite{yao2024taubench} adds simulated users and domain policies, and OSWorld~\cite{xie2024osworld} tests real desktops.
These benchmarks primarily evaluate successful completion rather than conditional action selection under an explicit price schedule.
EcoAgent-Bench differs by making cost a first-class task constraint: the same task can require different tool strategies depending on evidence quality and remaining budget.

\textbf{Cost-aware evaluation.}
CostBench~\cite{costbench2025} is the closest prior work: a cost-optimal multi-turn tool-planning benchmark with priced atomic and composite tools and dynamic cost changes.
EcoAgent-Bench is complementary rather than a correction of it: we emphasize \emph{real-derived} tasks, \emph{evidence-conditioned} escalation, explicit \emph{model-tier routing} and \emph{stop-loss} decisions, and a frozen agent/evaluator separation with anti-leakage controls.
FrugalGPT~\cite{chen2023frugalgpt} and RouteLLM~\cite{ong2024routellm} address cost via model cascading at the per-query level, not multi-step tool planning.
Kapoor et al.~\cite{kapoor2024aiagents} argue that agent evaluation must measure deployment-relevant factors including cost; we operationalize that argument.

\textbf{Tool use and abstention.}
Toolformer~\cite{schick2023toolformer}, Gorilla~\cite{patil2023gorilla}, and the survey of Qin et al.~\cite{qin2024toollearning} establish the tool-use landscape; SWE-agent~\cite{yang2024sweagent} and OpenHands~\cite{wang2024opendevin} provide workspace interfaces.
Our seed sources include HotpotQA~\cite{yang2018hotpotqa} and MuSiQue~\cite{trivedi2022musique} multi-hop QA.
The stop-loss family is grounded in the abstention tradition of SQuAD~2.0~\cite{rajpurkar2018know}: a competent system must decline to answer when no answer is supported.
LLM-as-a-judge scoring~\cite{zheng2023judging,liu2023geval} underpins our semantic evaluation.

\section{Benchmark Design}

\subsection{Task Structure}
Each task contains an instruction, a frozen priced environment visible to the agent, and oracle material visible only to the evaluator. It also specifies a \emph{canonical} budget-compliant trajectory, a contrasting \emph{high-regret} trajectory, and tight/medium/loose budgets derived from the canonical cost.
The canonical trace represents the task constructor's intended decision: it is usually cheaper, but must-upgrade items require the more expensive model tier. The contrasting trace captures either unnecessary escalation or a cheaper action that cannot solve the task.
Table~\ref{tab:costs} lists the shared retrieval actions; the supplementary material gives complete per-family accounting and the full registry.

\begin{table}[t]
\centering
\caption{Shared retrieval actions and abstract prices. The gap between cheap search and agentic deep research makes over-escalation measurable.}
\label{tab:costs}
\small
\setlength{\tabcolsep}{3pt}
\begin{tabular*}{\columnwidth}{@{\extracolsep{\fill}}lrl@{}}
\toprule
Tool & Cost & Tier / role \\
\midrule
\texttt{inspect\_tool}          & 4   & interface / discovery \\
\texttt{search\_tools}          & 8   & interface / discovery \\
\texttt{local\_keyword\_search} & 18  & atomic / cheap search \\
\texttt{read\_document}         & 22  & atomic / read a hit \\
\texttt{web\_search\_snapshot}  & 70  & medium / broader search \\
\texttt{deep\_research\_agent}  & 260 & composite / escalate \\
\texttt{page\_human\_oncall}    & 520 & human / last resort \\
\bottomrule
\end{tabular*}
\end{table}

\subsection{Economic Decisions}
The five families instantiate four economic decisions:
\begin{itemize}
\item \textbf{Do-not-over-escalate.} Cheap evidence already suffices; paying for a premium tool is wasteful. Carried by \texttt{cheap\_qa} (a low-cost search already surfaces the answer vs.\ agentic deep research).
\item \textbf{Escalation-required.} A single cheap search returns partial or misleading evidence; the answer lives at a source reachable only by iterative deep research. Carried by \texttt{escalation\_qa}.
\item \textbf{Model-tier routing.} All evidence is locally complete; the only decision is whether the reasoning demands an expensive frontier model or a cheap small model already suffices. Carried by \texttt{model\_upgrade\_qa} and the real-GAIA \texttt{frozen\_information\_qa} anchors.
\item \textbf{Stop-loss.} The premise is unsupported. The agent should gather evidence and abstain rather than continue spending or fabricate an answer. Carried by \texttt{stop\_loss}.
\end{itemize}

\subsection{Economic Rationality}
Given a task $t$ with budget $B$, an agent selects a trajectory $\pi=(a_1,\dots,a_k)$ with per-action costs $c(a_i)$ and total cost $C(\pi)=\sum_i c(a_i)$.
Let $\mathbb{1}[\pi\models t]$ indicate a correct answer.
The utility of a trajectory is
\[
U(\pi,t)=\begin{cases}
R-\lambda\,C(\pi) & \text{if } \mathbb{1}[\pi\models t]=1,\ C(\pi)\le B\\
-\lambda\,C(\pi)  & \text{if } \mathbb{1}[\pi\models t]=0,\ C(\pi)\le B\\
-\infty           & \text{if } C(\pi)>B,
\end{cases}
\]
where $R$ rewards correctness and $\lambda>0$ weights cost.
An economically rational agent maximizes $\mathbb{E}[U(\pi,t)]$: use cheap tools when they suffice, escalate or route up only when they do not, stop when nothing helps, and never exceed budget.

\subsection{Cost Model}
We use abstract units because provider prices and tool implementations change over time. The schedule preserves a coarse tier structure: bounded local operations are the base; broader retrieval costs several local operations; composite agents cost roughly an order of magnitude more; and human escalation is highest. Model-tier prices are defined on a separate 1:8 scale. Public search and model prices motivate these tiers~\cite{google_cse,gemini_grounding,perplexity_sonar}, but the units are not dollar estimates. The supplementary material reports the complete tier calibration and tool registry.

The scale is intentionally compressed so that escalation remains feasible on some tasks rather than being ruled out by construction. To test sensitivity to the chosen integers, we independently perturb every action cost by a factor drawn from $U(1/R,R)$. Across 50{,}000 draws for each of the seven family/decision strata, the ordering between canonical and high-regret trajectories is preserved in 100\% of draws for $R=2$ and at least 99.90\% for $R=3$. This analysis supports claims about relative action cost, not transfer to a particular deployment's billing model.

\section{Task Families}

\begin{table}[t]
\centering
\caption{Evaluation bundle: 304 tasks across five real-derived families.}
\label{tab:families}
\small
\setlength{\tabcolsep}{3pt}
\begin{tabular*}{\columnwidth}{@{\extracolsep{\fill}}p{0.46\columnwidth}r p{0.36\columnwidth}@{}}
\toprule
Family & $n$ & Required decision \\
\midrule
Escalation QA         & 115 & escalate retrieval \\
Model-upgrade QA      & 100 & route model tier \\
Cheap QA              & 72  & avoid QA escalation \\
Stop loss             & 10  & abstain after evidence \\
Frozen-information QA & 7   & route model tier (GAIA) \\
\midrule
\textbf{Total}                    & \textbf{304} & \\
\bottomrule
\end{tabular*}
\end{table}

\textbf{\texttt{frozen\_information\_qa} (7).}
Real GAIA 2023 validation questions with attachments frozen as local text, including spreadsheets whose cell background colors are preserved so that color/spatial questions remain grounded.
Because the evidence is locally complete, the decision is a model-tier one (cheap vs.\ frontier); these serve as real-GAIA anchors on the model-upgrade axis.

\textbf{\texttt{escalation\_qa} (115).}
Real-derived from HotpotQA multi-hop questions (96), GAIA tables (4), and MuSiQue answerable items (15).
One supporting hop is withheld to an authoritative source reachable only by iterative deep research, so the local (distractor-heavy) results are genuinely insufficient and answering \emph{requires} escalation.

\textbf{\texttt{cheap\_qa} (72).}
Real-derived from MuSiQue two-hop answerable items whose answer appears directly in ordinary (frozen) search-engine results: a low-cost search plus a quick read is sufficient, so paying for agentic deep research is wasteful over-escalation.
This family is the QA counterpart of escalation: the agent sees a page of search results, and the correct decision is the \emph{opposite}---escalate only if the answer is not already present.

\textbf{\texttt{stop\_loss} (10).}
Verified false-premise items are authored over MuSiQue distractor corpora and independently checked as unanswerable.
The target behavior is evidence-backed abstention rather than repeated search or fabrication.

\textbf{\texttt{model\_upgrade\_qa} (100).}
Real-derived from HotpotQA with all supporting evidence kept in context, so the only decision is which model tier to run.
The family is balanced 50 must-upgrade / 50 cheap-sufficient, with per-item fast/pro solvability recorded once offline and shipped as a lookup table, so scoring needs no live model call at evaluation time.

\section{Transformation Protocol}

For every family the pipeline applies the same high-level stages:
(1)~acquire or ingest a real seed pool with source metadata;
(2)~select rows deterministically (salted stable hashing) from the candidate pool;
(3)~adapt seed rows into budget-conditioned tasks with frozen evidence;
(4)~split evaluator-only oracle material from the agent-visible view;
(5)~materialize fixtures and projection manifests for external review;
(6)~run leakage, traceability, budget, source, packaging, and reproducibility checks.
The protocol records both inherited seed content and EcoAgent-added surfaces, so the benchmark is not simply ``SWE-bench / GAIA / multi-hop QA with a cost column appended.'' The budgeted tool environment, canonical and high-regret trajectories, and redacted agent view are part of the transformed task.

\section{Validity Controls}

\textbf{No leakage.}
In the agent view, expected answers appear only when they occur naturally in frozen evidence; evaluator rationales, tempting answers, tier labels, and reference trajectories are removed.

\textbf{Answer availability.}
For solvable tasks, the expected answer is recoverable from frozen evidence or an evaluator-supplied support field. The current episode-level grounding check is deliberately coarse: for several QA families it verifies evidence access, not entailment from the cited span.

\textbf{Escalation verification.}
Each \texttt{escalation\_qa} task is checked against four conditions: cheap search cannot reach the authoritative source, the answer is absent from every cheap-visible document, deep research reaches it, and a scripted cheap agent fails while an escalating agent solves it.

\textbf{Model-tier verification.}
Cheap-sufficient items require a small model to answer correctly on all three attempts; must-upgrade items require the small model to fail all three while a frontier model succeeds on a majority---so labels reflect a robust capability gap rather than single-shot noise.
Budget is uniform across the family so its magnitude does not leak the correct tier, while a cost-regret term still penalizes over-escalation.

\textbf{Stop-loss verification.}
Stop tasks are independently checked as false-premise items. The scorer requires an insufficient-evidence abstention after evidence access; it does not require the canonical deep-research call.

\textbf{Difficulty cues.}
Task IDs are hashed in workspace runs, budgets are shared within each family, and reference costs are hidden. Workspace templates still describe the tools available to a family and may therefore cue the action class; we treat workspace results as interface-level measurements rather than clean estimates of prompt-independent routing.

\textbf{Regression evidence.}
Expected-fail smoke tests inject each bad state (leaked answers, tampered result bindings, over-broad redistribution) and verify that the corresponding gate rejects it, giving evidence that the checks are not only happy-path reports.

\section{Reproducibility and Packaging}

The bundle is reproduced by a single wrapper command that rebuilds every family from its seed pool, re-runs all gates, and regenerates the audit packet, reproducibility capsule, and release manifest.
The wrapper separates machine-checkable gates (leakage, provenance, packaging, and reproducibility) from the human realism audit and reports their status independently.

\textbf{Audit packet.}
Human realism review covers 45 final-bundle tasks (14.8\%): all seven frozen-information and all ten stop-loss tasks, together with 13 escalation-QA, 11 model-upgrade, and four cheap-QA tasks. The review interface presents each family with its native evidence surface; for \texttt{cheap\_qa}, the frozen corpus appears as ordinary search results, consistent with the ``a plain search already surfaces the answer'' framing.

\textbf{Result artifacts.}
Every reported episode is bound to the 304-task bundle by task-file SHA256, canonical task SHA256, task IDs, family counts, agent names, and budget labels, so the leaderboard (\S Results) is reproducible against the frozen release.
The judge re-scoring of QA answers is performed offline over stored agent outputs, requiring no agent re-execution.

\section{Experimental Setup}

We evaluate \textbf{seven LLM agents and four scripted controls across three tracks} on all 304 tasks.
\emph{Scripted} (4): CheapFirst, CopilotFirst, RetryLoop, and a BudgetAwarePlanner, in oracle composite mode. These are \textbf{construct-validity controls, not information-fair competitors}: they run with oracle composite tools and read the gold answer once their policy reaches an answer-bearing step, so their budgeted accuracy measures only whether the policy's tool/stop choices stay within budget---not answer derivation. They are therefore not directly comparable to the evidence-only LLM tracks, and we do not rank them against LLM agents.
\emph{Tool-API LLM} (3): a multi-turn tool-calling loop (up to 10 turns) over the priced tools, with Sonnet, GPT-5.4, and Gemini~2.5~Pro backbones (evidence-only composite mode). The reported results use the balanced prompt, one episode per task ($n{=}304$ per backbone).
\emph{Workspace CLI} (4): Claude Code (Opus~4.8 and Haiku backbones) and Codex (GPT-5.5 and GPT-5.4-mini) in materialized workspaces.
Open-ended QA is graded by \texttt{claude-sonnet-4-6} at temperature zero using a fixed binary rubric prompt reproduced in the supplementary material; \texttt{stop\_loss} requires an insufficient-evidence abstention. One human reviewer re-labeled 96 responses, sampled evenly across four QA families and the primary judge's positive/negative decisions in an interface that hid model and judge fields by default. Human--judge agreement is 91.7\% with Cohen's $\kappa=0.833$; all eight disagreements are judge false positives. The supplementary material reports reweighted estimates and model-level checks.
\textbf{Strict budgeted success} requires a correct response, the applicable evidence-access check, and no budget violation.

The scripted and tool-API tracks use the shared action ledger. The workspace harness does not recover the agent-visible priced action sequence. It instead records a post-hoc execution proxy,
$10+0.5T+10M+5N$, from wall time $T$ (seconds), modified files $M$, and new files $N$. This quantity was not visible to the agent and is not a measure of economical tool choice; we report it only as a systems-effort diagnostic. All workspace runs remain below the proxy cap, so their reported success equals correctness. We draw cost-aware conclusions only from the scripted and tool-API tracks.

To test whether agents \emph{condition} on the budget (not merely act under one fixed budget), we additionally sweep GPT-5.4 over three budgets per family, calibrated so that tight sits \emph{below} the escalation-path cost and medium/loose sit \emph{above} it (\S Finding 5).

\section{Results}

Table~\ref{tab:casestudy} shows two tasks with opposite decisions: one requires escalation, whereas the other is solved by local evidence. Neither extreme scripted policy succeeds on both; GPT-5.4 also stops short of the required escalation in the first case.

\begin{table*}[t]
\centering
\small
\renewcommand{\arraystretch}{1.10}
\caption{Two real tasks with opposite optimal actions. The scripted controls have oracle answer access; GPT-5.4-neutral is evidence-only. $\checkmark$/$\times$ denote strict budgeted success.}
\label{tab:casestudy}
\begin{tabular}{@{}l p{0.40\textwidth} p{0.38\textwidth}@{}}
\toprule
 & \textbf{Escalation required} ($B{=}480$) & \textbf{Cheap evidence sufficient} ($B{=}90$) \\
 & \emph{1977 album song references which comics brand?} $\rightarrow$ Marvel & \emph{Who introduced the 14th-c.\ notation system?} $\rightarrow$ John Kukuzelis \\
\midrule
Evidence structure & Local evidence is insufficient & Local document contains the answer \\
\addlinespace[2pt]
CheapFirst & local search $\rightarrow$ abstain; $C{=}82$; strict~$\times$ & ``John Kukuzelis''; $C{=}82$; strict~$\checkmark$ \\
\addlinespace[2pt]
CopilotFirst & deep research $\rightarrow$ ``Marvel''; $C{=}322$; strict~$\checkmark$ & correct answer, but $C{=}364{>}B$; strict~$\times$ \\
\addlinespace[2pt]
GPT-5.4-neutral & local $+$ web search $\rightarrow$ abstain; $C{=}228$; strict~$\times$ & --- (not evaluated) \\
\bottomrule
\end{tabular}
\end{table*}

\begin{table}[t]
\centering
\caption{Results on the 304-task bundle. \emph{Micro} is task-averaged strict budgeted success. \emph{Up}/\emph{Save} are strict accuracy on the upgrade-oriented family group (escalation, model-upgrade, frozen; 222 tasks) and save-oriented group (cheap-QA, stop-loss; 82 tasks); model-tier families contain both cheap-sufficient and must-upgrade items. \emph{Econ}${=}\min(\text{Up},\text{Save})$. Tool-API rows use the balanced prompt (one episode per task). Units are ledger cost for scripted/tool-API rows and a post-hoc execution proxy for workspace rows. Rows are ordered by Econ within each track.}
\label{tab:leaderboard}
\small
\renewcommand{\arraystretch}{0.94}
\setlength{\tabcolsep}{3pt}
\begin{tabular*}{\columnwidth}{@{\extracolsep{\fill}}lrrrrr@{}}
\toprule
Agent & Econ & Up & Save & Micro & Units \\
\midrule
\multicolumn{6}{@{}l}{\emph{Workspace CLI}} \\
Claude Code (Opus 4.8) & \textbf{53.6\%} & 54\% & 96\% & 65.1\% & 37 \\
Codex (GPT-5.5)        & 49.5\% & 50\% & 74\% & 56.2\% & 51 \\
Codex (GPT-5.4-mini)   & 46.4\% & 46\% & 74\% & 53.9\% & 50 \\
Claude Code (Haiku)    & 44.6\% & 45\% & 90\% & 56.9\% & 37 \\
\addlinespace[2pt]
\multicolumn{6}{@{}l}{\emph{Scripted controls (oracle)}} \\
CheapFirst             & 23.9\% & 24\%  & 34\% & 26.6\% & 71 \\
BudgetAware            & 22.0\% & 50\%  & 22\% & 42.4\% & 173 \\
CopilotFirst           & 12.2\% & 100\% & 12\% & 76.3\% & 335 \\
RetryLoop              & 0.0\%  & 100\% & 0\%  & 73.0\% & 441 \\
\addlinespace[2pt]
\multicolumn{6}{@{}l}{\emph{Tool-API LLM}} \\
Sonnet         & 7.3\% & 30\% & 7\% & 24.0\% & 207 \\
Gemini 2.5 Pro & 6.1\% & 22\% & 6\% & 17.4\% & 268 \\
GPT-5.4        & 3.6\% & 4\%  & 5\% & 3.9\%  & 97 \\
\bottomrule
\end{tabular*}
\end{table}

\begin{figure*}[t]
\centering
\includegraphics[width=\textwidth]{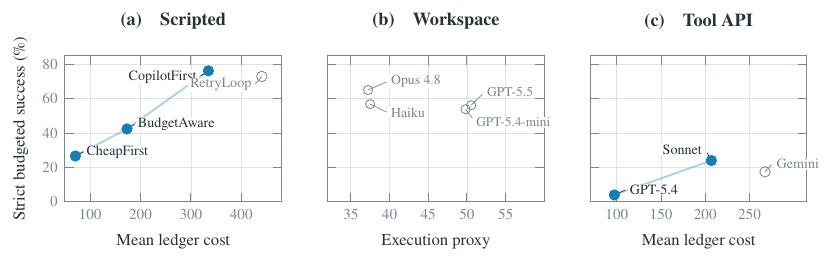}
\caption{Success versus resource use. Panels (a,c) use the shared action ledger; filled points mark their Pareto frontiers. Panel (b) uses a post-hoc runtime/edit proxy, shown only as a systems diagnostic. Units are not comparable across panels.}
\label{fig:pareto}
\end{figure*}

\subsection{Finding 1: Micro-Accuracy Rewards One-Sided Policies}
Plain micro-averaged strict success assigns the two always-escalate oracle controls 76.3\% and 73.0\% (Table~\ref{tab:leaderboard}) even though they spend the most and score only 12.2\% and 0.0\% on the save-oriented group. The 222-task Up group should not be read as 222 items that all require escalation: its two model-tier families deliberately contain both cheap-sufficient and must-upgrade items.
The economic-consistency score $\text{Econ}=\min(\text{Up},\text{Save})$ makes the controls' one-sided behavior explicit by retaining their worse group score. It is a regime-balanced diagnostic rather than a claim that the oracle controls and LLM agents are information-fair competitors. Among the seven LLM agents, the four workspace CLI configurations attain 44.6--53.6\% Econ, whereas the tool-API agents attain 3.6--7.3\%. Claude Code (Opus~4.8) has the highest reported LLM score, with 54\% on Up and 96\% on Save.
To separate completion from expenditure at the dataset level we also report $S_\lambda=s-\lambda C/C_{\max}$, where $s$ is micro strict success and $C_{\max}$ is the largest mean cost among the scripted controls; the full sweep appears in the supplementary material.

\subsection{Finding 2: Tool-API Agents Under-Escalate and Abandon}
Every tool-API agent scores at most 24.0\% micro strict success and at most 7.3\% economic consistency, driven by two failure modes: over-exploration that busts the budget (Sonnet and Gemini bust it on 74\% and 99\% of cheap-QA tasks) and ungrounded prior-knowledge guessing (GPT-5.4 answers cheap-QA at low cost but only 25\% lenient accuracy). Across the 294 non-stop QA tasks, Gemini returns an insufficient-evidence abstention in 173 episodes (59\%) and GPT-5.4 in 137 (47\%). On escalation-required QA specifically, 45/115 GPT-5.4 episodes abstain and none invokes deep research. GPT-5.4 therefore has the lowest mean cost in this track but the highest cost per success.

\begin{figure*}[t]
\centering
\includegraphics[width=0.92\textwidth]{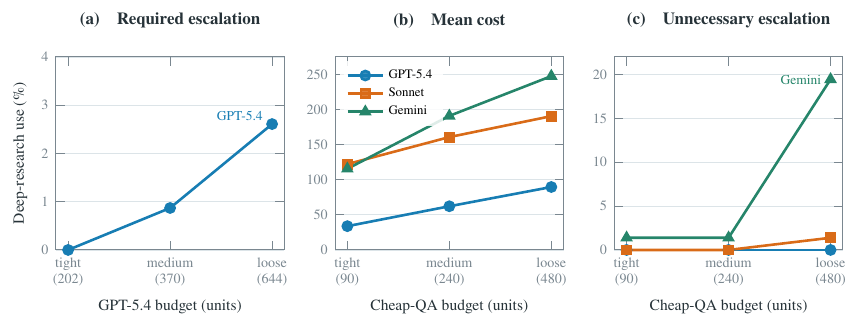}
\caption{Budget conditioning exposes opposite errors. (a) On 115 tasks that require deep research, GPT-5.4 rarely escalates even after the budget crosses the $\approx$322 path cost. (b,c) On 72 tasks already solvable with cheap evidence, all three backbones spend more as the budget grows; Gemini additionally raises unnecessary deep-research use to 19.4\% at the loose budget. Budget grids are family-specific and printed on each panel.}
\label{fig:budgetcond}
\end{figure*}

\subsection{Finding 3: Opus Leads the Workspace Track}
Claude Code with an Opus~4.8 backbone achieves the highest workspace micro success, 65.1\%, and the highest economic-consistency score among the evaluated LLM agents (53.6\%). The stronger backbone improves micro success within both CLI systems (Opus over Haiku; GPT-5.5 over GPT-5.4-mini). We do not interpret the workspace execution proxy as tool-use economy: it is driven partly by runtime, is not the price schedule shown to the agent, and differs from the tool-API ledger.

\subsection{Finding 4: Scripted Controls Validate the Decision Structure}
The scripted policies read the gold answer (oracle controls), so we do \emph{not} compare them to the LLM agents; they instead verify that the benchmark's intended structure is realizable. A budget-aware policy stays within budget on every family (0 violations), the always-escalate policies bust the budget on exactly the do-not-over-escalate families (CopilotFirst scores 12\% and RetryLoop 0\% on the save regime), and CheapFirst fails the escalation-required ones. This confirms that the five families encode genuinely opposite optimal actions that no single fixed policy satisfies---the property the economic-consistency score is designed to reward and the LLM tracks are then measured against.

\textbf{Bundle balance.} Among the 294 answerable QA tasks, 169 favor a costlier action (115 retrieval escalations, 50 model upgrades, and four frozen-information model upgrades), while 125 favor a cheaper action (72 cheap-QA, 50 cheap-sufficient model-tier, and three cheap-sufficient frozen-information tasks). The remaining ten tasks require stopping after evidence collection. Thus the answerable subset is 57.5\%/42.5\%, not 73\%/27\%; the 222/82 split in Table~\ref{tab:leaderboard} is a grouping of \emph{families}, not a count of optimal action directions.

\subsection{Finding 5: Budget Sensitivity Is Weak or Misdirected}
We first sweep GPT-5.4 over tight (202), medium (370), and loose (644) budgets; the latter two exceed the $\approx$322 escalation-path cost. On \texttt{escalation\_qa}, deep-research use changes only from 0\% to 0.9\% to 2.6\% (0/115, 1/115, and 3/115; Figure~\ref{fig:budgetcond}), while 64--79\% of episodes make no tool call.

We next evaluate all three backbones on the expanded 72-task \texttt{cheap\_qa} sweep. Its family-specific tight/medium/loose budgets are 90/240/480. Mean cost rises monotonically: 34/62/90 units for GPT-5.4, 122/161/191 for Sonnet, and 116/191/248 for Gemini. Yet this extra budget does not improve action selection. GPT-5.4 never invokes deep research, Sonnet does so once at the loose budget (1.4\%), and Gemini rises from 1.4\% to 1.4\% to 19.4\%---on tasks where cheap evidence already suffices. Thus the clearest budget response is asymmetric and misdirected: GPT-5.4 barely buys the required escalation, whereas Gemini uses a loose budget to buy unnecessary escalation.

\section{Discussion}

\textbf{Budget compliance is not budget conditioning.}
An agent can remain under budget without using the budget to guide its decisions. GPT-5.4 illustrates this distinction: it is inexpensive largely because it stops or answers early, not because it reliably identifies when cheap evidence is sufficient. Conversely, Gemini's higher loose-budget expenditure on cheap-QA tasks shows that spending more when more is available is not, by itself, adaptive behavior. A useful policy therefore needs two signals that current agents often conflate: whether the evidence is sufficient and whether the next action is affordable.

\textbf{The benchmark localizes different failure sources.}
Because each episode records tool calls, evidence access, correctness, and budget feasibility, a failure can be attributed to one of three stages. A \emph{selection error} chooses the wrong action class, such as stopping before required escalation or escalating after the answer is already visible. An \emph{execution error} chooses a viable path but still produces the wrong answer. A \emph{budget error} reaches a correct answer only after exceeding the task budget. This distinction matters for agent development: better retrieval cannot repair a policy that never invokes it, while a better router cannot repair an evaluator or tool that fails after the correct route is chosen.

\textbf{Evaluation should preserve the decision structure.}
Micro accuracy alone rewards the majority regime and can make a fixed always-escalate policy appear strong. Mean cost alone similarly favors agents that terminate prematurely. We therefore recommend reporting a family-level success matrix together with feasibility and cost, and using Econ only as a compact worst-regime diagnostic. The full matrix remains necessary: the same Econ value can arise from different mixtures of under-escalation, over-escalation, and answer-generation failure. Comparisons should also remain within interfaces that expose the same actions and accounting rules; the workspace results are informative about completion across decision regimes, but not about adherence to the shared tool ledger.

\textbf{Paired regimes are diagnostic, not causal.}
The families are organized by the decision they test, rather than only by source dataset. Escalation-QA and cheap-QA present the same broad choice---whether to purchase deeper retrieval---under opposite evidence states. Model-upgrade QA holds the evidence interface and budget fixed while reversing which model tier is sufficient. Stop-loss adds abstention, so economical behavior cannot be reduced to always choosing the cheapest answering path. These are regime-level controls, not item-level counterfactual pairs: questions and sources differ across families, and a score difference between families cannot be attributed to one isolated intervention. Their role is diagnostic. A policy that succeeds only when the optimal action is consistently expensive or consistently cheap is exposed by the opposite regime; a conditional policy must succeed on both.

\textbf{Implications for agent design.}
The results point to a controller that estimates the marginal value of the next action, rather than following a fixed escalation ladder. Such a controller should first ask whether the current evidence supports an answer; if not, it should compare the expected gain from each available action with both its price and the remaining budget. This separates three decisions that monolithic prompting often collapses: whether to continue, which resource to invoke, and whether the resulting evidence is sufficient to stop. EcoAgent-Bench does not prescribe how to implement that controller. It provides paired regimes in which each decision can be tested: cheap-QA and stop-loss penalize unnecessary continuation, escalation-QA rewards warranted retrieval, and model-tier tasks test whether stronger inference is purchased only when needed. The weak and sometimes reversed response in the budget sweeps suggests that merely exposing a budget in the prompt is insufficient; budget awareness must enter the action-selection rule itself.

\section{Limitations}

Human realism review covers 45 of 304 tasks rather than the full bundle. Main LLM runs are single-shot at temperature zero. Judge calibration uses one human reviewer, so it estimates human--judge but not human--human agreement. Sonnet and GPT-5.4 answers have the same $+12.5$-point judge-minus-human positive-rate gap in this sample, which does not indicate a uniquely Sonnet-specific bias but is not an equivalence test. For several QA families, ``grounded'' means that the episode accessed evidence, not that a cited span entails the answer. Stop-loss scoring likewise accepts evidence-backed abstention without requiring the canonical deep-research confirmation.

Costs are abstractions. Tool-API costs follow the fixed action ledger. Workspace agents, however, saw a separate file-operation price list (including a 200-unit research request), whereas the reported execution proxy was computed after the run from latency and file changes; it neither implements the shared 260-unit research price nor reveals which priced action was chosen. We therefore make no cross-track cost, Pareto, or tool-economy claim from workspace runs; their Econ score is only a regime-balanced completion diagnostic. Workspace templates also differ by family and may reveal the available escalation interface. Among answerable tasks, costlier actions are optimal for 57.5\% and cheaper actions for 42.5\%; the ten stop-loss items form a separate abstention decision. The main tool-API leaderboard uses one balanced-prompt episode per task, while the budget sweeps use one episode at each family-specific budget. These choices limit claims about absolute deployment cost, prompt robustness, and cross-interface comparisons.

\section{Conclusion}

EcoAgent-Bench evaluates whether an agent's tool and model choices fit both the evidence state and the available budget. Across 304 real-derived tasks, strict success and trajectory cost expose different failure modes: aggressive controls overspend on save-oriented regimes, whereas tool-API agents often stop before warranted escalation. The threshold-crossing sweep further shows that more budget does not reliably improve decisions: GPT-5.4 rarely purchases required retrieval, while Gemini increasingly purchases it when cheap evidence already suffices.

The central result is not simply that current agents are costly or inaccurate, but that completion, budget feasibility, and conditional resource selection are distinct capabilities. Low expenditure can reflect premature stopping, just as high completion can conceal indiscriminate escalation. Family-level reporting and economic consistency expose these one-sided policies; released traces and integrity-bound artifacts support action-level diagnosis. EcoAgent-Bench thus provides a testbed for controllers that jointly estimate evidence sufficiency and the value of the next action, turning a stated budget from a passive limit into part of agent reasoning.

\bibliography{aaai2026}

@inproceedings{mialon2023gaia,
  title={GAIA: A Benchmark for General AI Assistants},
  author={Mialon, Gr{\'e}goire and Fourrier, Cl{\'e}mentine and Swift, Craig and Wolf, Thomas and LeCun, Yann and Scialom, Thomas},
  booktitle={The Twelfth International Conference on Learning Representations},
  url={https://openreview.net/forum?id=fibxvahvs3},
  year={2024}
}

@inproceedings{jimenez2023swebench,
  title={SWE-bench: Can Language Models Resolve Real-World GitHub Issues?},
  author={Jimenez, Carlos E. and Yang, John and Wettig, Alexander and Yao, Shunyu and Pei, Kexin and Press, Ofir and Narasimhan, Karthik},
  booktitle={The Twelfth International Conference on Learning Representations},
  url={https://openreview.net/forum?id=VTF8yNQM66},
  year={2024}
}

@inproceedings{zhou2023webarena,
  title={WebArena: A Realistic Web Environment for Building Autonomous Agents},
  author={Zhou, Shuyan and Xu, Frank F. and Zhu, Hao and Zhou, Xuhui and Lo, Robert and Sridhar, Abishek and Cheng, Xianyi and Ou, Tianyue and Bisk, Yonatan and Fried, Daniel and Alon, Uri and Neubig, Graham},
  booktitle={The Twelfth International Conference on Learning Representations},
  url={https://openreview.net/forum?id=rmiwIL98uQ},
  year={2024}
}

@inproceedings{liu2023agentbench,
  title={AgentBench: Evaluating LLMs as Agents},
  author={Liu, Xiao and Yu, Hao and Zhang, Hanchen and Xu, Yifan and Lei, Xuanyu and Lai, Hanyu and Gu, Yu and Ding, Hangliang and Men, Kaiwen and Yang, Kejuan and Zhang, Shudan and Deng, Xiang and Zeng, Aohan and Du, Zhengxiao and Zhang, Chenhui and Shen, Sheng and Zhang, Tianjun and Su, Yu and Sun, Huan and Huang, Minlie and Dong, Yuxiao and Tang, Jie},
  booktitle={The Twelfth International Conference on Learning Representations},
  url={https://openreview.net/forum?id=zAdUB0aCTQ},
  year={2024}
}

@inproceedings{yao2024taubench,
  title={$\tau$-bench: A Benchmark for Tool-Agent-User Interaction in Real-World Domains},
  author={Yao, Shunyu and Shinn, Noah and Razavi, Pedram and Narasimhan, Karthik},
  booktitle={The Thirteenth International Conference on Learning Representations},
  url={https://openreview.net/forum?id=roNSXZpUDN},
  year={2025}
}

@article{xie2024osworld,
  title={OSWorld: Benchmarking Multimodal Agents for Open-Ended Tasks in Real Computer Environments},
  author={Xie, Tianbao and Zhang, Danyang and Chen, Jixuan and Li, Xiaochuan and Zhao, Siheng and Cao, Ruisheng and Hua, Toh Jing and Cheng, Zhoujun and Shin, Dongchan and Lei, Fangyu and Liu, Yitao and Xu, Yiheng and Zhou, Shuyan and Savarese, Silvio and Xiong, Caiming and Zhong, Victor and Yu, Tao},
  journal={Advances in Neural Information Processing Systems},
  volume={37},
  doi={10.52202/079017-1650},
  url={https://proceedings.neurips.cc/paper_files/paper/2024/hash/5d413e48f84dc61244b6be550f1cd8f5-Abstract-Datasets_and_Benchmarks_Track.html},
  year={2024}
}

@article{costbench2025,
  title={CostBench: Evaluating Multi-Turn Cost-Optimal Planning and Adaptation in Dynamic Environments for LLM Tool-Use Agents},
  author={Liu, Jiayu and Qian, Cheng and Su, Zhaochen and Zong, Qing and Huang, Shijue and He, Bingxiang and Fung, Yi R.},
  journal={arXiv preprint arXiv:2511.02734},
  doi={10.48550/arXiv.2511.02734},
  url={https://arxiv.org/abs/2511.02734},
  year={2025}
}

@article{chen2023frugalgpt,
  title={FrugalGPT: How to Use Large Language Models While Reducing Cost and Improving Performance},
  author={Chen, Lingjiao and Zaharia, Matei and Zou, James},
  journal={Transactions on Machine Learning Research},
  url={https://openreview.net/forum?id=cSimKw5p6R},
  year={2024}
}

@article{ong2024routellm,
  title={RouteLLM: Learning to Route LLMs with Preference Data},
  author={Ong, Isaac and Almahairi, Amjad and Wu, Vincent and Chiang, Wei-Lin and Wu, Tianhao and Gonzalez, Joseph E. and Kadous, M Waleed and Stoica, Ion},
  journal={arXiv preprint arXiv:2406.18665},
  doi={10.48550/arXiv.2406.18665},
  url={https://arxiv.org/abs/2406.18665},
  year={2024}
}

@article{kapoor2024aiagents,
  title={AI Agents That Matter},
  author={Kapoor, Sayash and Stroebl, Benedikt and Siegel, Zachary S. and Nadgir, Nitya and Narayanan, Arvind},
  journal={arXiv preprint arXiv:2407.01502},
  doi={10.48550/arXiv.2407.01502},
  url={https://arxiv.org/abs/2407.01502},
  year={2024}
}

@article{schick2023toolformer,
  title={Toolformer: Language Models Can Teach Themselves to Use Tools},
  author={Schick, Timo and Dwivedi-Yu, Jane and Dess{\`i}, Roberto and Raileanu, Roberta and Lomeli, Maria and Hambro, Eric and Zettlemoyer, Luke and Cancedda, Nicola and Scialom, Thomas},
  journal={Advances in Neural Information Processing Systems},
  volume={36},
  doi={10.52202/075280-2997},
  url={https://proceedings.neurips.cc/paper_files/paper/2023/hash/d842425e4bf79ba039352da0f658a906-Abstract-Conference.html},
  year={2023}
}

@article{patil2023gorilla,
  title={Gorilla: Large Language Model Connected with Massive APIs},
  author={Patil, Shishir G. and Zhang, Tianjun and Wang, Xin and Gonzalez, Joseph E.},
  journal={Advances in Neural Information Processing Systems},
  volume={37},
  doi={10.52202/079017-4020},
  url={https://proceedings.neurips.cc/paper_files/paper/2024/hash/e4c61f578ff07830f5c37378dd3ecb0d-Abstract-Conference.html},
  year={2024}
}

@article{qin2024toollearning,
  title={Tool Learning with Foundation Models},
  author={Qin, Yujia and Hu, Shengding and Lin, Yankai and Chen, Weize and Ding, Ning and Cui, Ganqu and Zeng, Zheni and Zhou, Xuanhe and Huang, Yufei and Xiao, Chaojun and Han, Chi and Fung, Yi Ren and Su, Yusheng and Wang, Huadong and Qian, Cheng and Tian, Runchu and Zhu, Kunlun and Liang, Shihao and Shen, Xingyu and Xu, Bokai and Zhang, Zhen and Ye, Yining and Li, Bowen and Tang, Ziwei and Yi, Jing and Zhu, Yuzhang and Dai, Zhenning and Yan, Lan and Cong, Xin and Lu, Yaxi and Zhao, Weilin and Huang, Yuxiang and Yan, Junxi and Han, Xu and Sun, Xian and Li, Dahai and Phang, Jason and Yang, Cheng and Wu, Tongshuang and Ji, Heng and Li, Guoliang and Liu, Zhiyuan and Sun, Maosong},
  journal={ACM Computing Surveys},
  volume={57},
  number={4},
  pages={101:1--101:40},
  doi={10.1145/3704435},
  url={https://doi.org/10.1145/3704435},
  year={2024}
}

@article{yang2024sweagent,
  title={SWE-agent: Agent-Computer Interfaces Enable Automated Software Engineering},
  author={Yang, John and Jimenez, Carlos E. and Wettig, Alexander and Lieret, Kilian and Yao, Shunyu and Narasimhan, Karthik and Press, Ofir},
  journal={Advances in Neural Information Processing Systems},
  volume={37},
  pages={50528--50652},
  doi={10.52202/079017-1601},
  url={https://proceedings.neurips.cc/paper_files/paper/2024/hash/5a7c947568c1b1328ccc5230172e1e7c-Abstract-Conference.html},
  year={2024}
}

@inproceedings{wang2024opendevin,
  title={OpenHands: An Open Platform for AI Software Developers as Generalist Agents},
  author={Wang, Xingyao and Li, Boxuan and Song, Yufan and Xu, Frank F. and Tang, Xiangru and Zhuge, Mingchen and Pan, Jiayi and Song, Yueqi and Li, Bowen and Singh, Jaskirat and Tran, Hoang H. and Li, Fuqiang and Ma, Ren and Zheng, Mingzhang and Qian, Bill and Shao, Yanjun and Muennighoff, Niklas and Zhang, Yizhe and Hui, Binyuan and Lin, Junyang and Brennan, Robert and Peng, Hao and Ji, Heng and Neubig, Graham},
  booktitle={The Thirteenth International Conference on Learning Representations},
  url={https://openreview.net/forum?id=OJd3ayDDoF},
  year={2025}
}

@article{zheng2023judging,
  title={Judging LLM-as-a-Judge with MT-Bench and Chatbot Arena},
  author={Zheng, Lianmin and Chiang, Wei-Lin and Sheng, Ying and Zhuang, Siyuan and Wu, Zhanghao and Zhuang, Yonghao and Lin, Zi and Li, Zhuohan and Li, Dacheng and Xing, Eric P. and Zhang, Hao and Gonzalez, Joseph E. and Stoica, Ion},
  journal={Advances in Neural Information Processing Systems},
  volume={36},
  doi={10.52202/075280-2020},
  url={https://proceedings.neurips.cc/paper_files/paper/2023/hash/91f18a1287b398d378ef22505bf41832-Abstract-Datasets_and_Benchmarks.html},
  year={2023}
}

@inproceedings{liu2023geval,
  title={G-Eval: NLG Evaluation using GPT-4 with Better Human Alignment},
  author={Liu, Yang and Iter, Dan and Xu, Yichong and Wang, Shuohang and Xu, Ruochen and Zhu, Chenguang},
  booktitle={Proceedings of the 2023 Conference on Empirical Methods in Natural Language Processing},
  pages={2511--2522},
  address={Singapore},
  publisher={Association for Computational Linguistics},
  doi={10.18653/v1/2023.emnlp-main.153},
  url={https://aclanthology.org/2023.emnlp-main.153/},
  year={2023}
}

@inproceedings{yang2018hotpotqa,
  title={{HotpotQA}: A Dataset for Diverse, Explainable Multi-hop Question Answering},
  author={Yang, Zhilin and Qi, Peng and Zhang, Saizheng and Bengio, Yoshua and Cohen, William W. and Salakhutdinov, Ruslan and Manning, Christopher D.},
  booktitle={Proceedings of the 2018 Conference on Empirical Methods in Natural Language Processing},
  pages={2369--2380},
  address={Brussels, Belgium},
  publisher={Association for Computational Linguistics},
  doi={10.18653/v1/D18-1259},
  url={https://aclanthology.org/D18-1259/},
  year={2018}
}

@article{trivedi2022musique,
  title={{MuSiQue}: Multihop Questions via Single-hop Question Composition},
  author={Trivedi, Harsh and Balasubramanian, Niranjan and Khot, Tushar and Sabharwal, Ashish},
  journal={Transactions of the Association for Computational Linguistics},
  volume={10},
  pages={539--554},
  doi={10.1162/tacl_a_00475},
  url={https://aclanthology.org/2022.tacl-1.31/},
  year={2022}
}

@inproceedings{rajpurkar2018know,
  title={Know What You Don't Know: Unanswerable Questions for {SQuAD}},
  author={Rajpurkar, Pranav and Jia, Robin and Liang, Percy},
  booktitle={Proceedings of the 56th Annual Meeting of the Association for Computational Linguistics (Volume 2: Short Papers)},
  pages={784--789},
  address={Melbourne, Australia},
  publisher={Association for Computational Linguistics},
  doi={10.18653/v1/P18-2124},
  url={https://aclanthology.org/P18-2124/},
  year={2018}
}

@misc{google_cse,
  title={{Custom Search JSON API}: Pricing},
  author={{Google}},
  year={2026},
  howpublished={\url{https://developers.google.com/custom-search/v1/overview}},
  note={Existing customers: 100 free queries/day, then \$5 per 1{,}000 queries. Representative rate, accessed July 2026}
}

@misc{gemini_grounding,
  title={{Gemini API}: Grounding with {Google} Search},
  author={{Google}},
  year={2026},
  howpublished={\url{https://ai.google.dev/gemini-api/docs/grounding}},
  note={\$14 per 1{,}000 grounded queries (current-generation). Representative rate, accessed July 2026}
}

@misc{perplexity_sonar,
  title={{Perplexity Sonar API}: Pricing},
  author={{Perplexity AI}},
  year={2026},
  howpublished={\url{https://docs.perplexity.ai/docs/getting-started/pricing}},
  note={Sonar request fee \$5--\$14 per 1{,}000 requests plus token cost (\$1--\$3/1M input, \$1--\$15/1M output). Representative rates, accessed July 2026}
}

\end{document}